%% file: main.tex
\documentclass{article}

\usepackage[final]{corl_2024}
\usepackage{times}

\usepackage[normalem]{ulem}

\usepackage[numbers]{natbib}
\usepackage{multicol}

\usepackage{times}
\usepackage{float}
\usepackage{booktabs}
\usepackage{graphicx}
\usepackage{hyperref}
\usepackage{multirow}
\usepackage{balance}
\usepackage{placeins}
\usepackage{soul}
\usepackage{standalone}
\usepackage[font={small}]{caption}
\usepackage{subcaption}
\usepackage{tikz}
\usepackage{url}
\usepackage{amsmath, amssymb}
\usepackage{dsfont}
\usepackage{mathtools}
\usepackage{algorithm}
\usepackage{algpseudocode}
\usepackage{xcolor}
\MakeRobust{\Call}

\algnewcommand{\Optional}{\item[\textbf{Optional:}]}

\usepackage{wrapfig}
\usepackage[normalem]{ulem}

\usepackage{colortbl}
\usepackage{framed}      
\usepackage{changepage}  

\newcommand{\ourmethod}{\textsc{ORION}}

\newcommand{\myparagraph}[1]{\noindent\textbf{#1}}

\definecolor{formalshade}{rgb}{0.85, 0.95, 0.95}
\newenvironment{formal}{%
  \MakeFramed{\advance\hsize-\width\FrameRestore}%
  \noindent\hspace{-4.55pt}%
  \begin{adjustwidth}{}{5pt}%
  \vspace{-7pt}\vspace{0.1pt}%
}
{%
  \vspace{0.1pt}\end{adjustwidth}\endMakeFramed%
}

\newcommand{\G}{\mathcal{G}}

\newcommand{\HandNode}{h}
\newcommand{\target}{\text{target}}
\newcommand{\reference}{\text{ref}}
\newcommand{\human}{V}
\newcommand{\robot}{Ro}

\newcommand{\mugcoaster}{\texttt{Mug-on-coaster}} 
\newcommand{\simpleboat}{\texttt{Simple-boat-assembly}}
\newcommand{\chip}{\texttt{Chips-on-plate}}
\newcommand{\llama}{\texttt{Succulents-in-llama-vase}}
\newcommand{\rearrange}{\texttt{Rearrange-mug-box}} 
\newcommand{\complexboat}{\texttt{Complex-boat-assembly}}
\newcommand{\breakfast}{\texttt{Prepare-breakfast}} 
\newcommand{\juice}{\texttt{Pour-juice}} 
\newcommand{\peasplate}{\texttt{Peas-on-plate}} 
\newcommand{\cheeseplate}{\texttt{Cheese-on-plate}} 

\DeclareMathOperator*{\argmin}{arg\,min}
\newcommand{\norm}[1]{\left\lVert#1\right\rVert}

\newcommand{\loosepar}{\looseness=-1}



\begin{document}


\title{Vision-based Manipulation from Single Human Video with Open-World Object Graphs}


\author{Yifeng Zhu$^{*,1}$, Arisrei Lim$^{*,1}$, Peter Stone$^{1,2}$, Yuke Zhu$^{1}$ \\ ~\\$^{1}$The University of Texas at Austin $^{2}$Sony AI\vspace{-0.8cm}}
\maketitle

\begin{abstract}
This work presents an object-centric approach to learning vision-based manipulation skills from human videos. We investigate the problem of robot manipulation via imitation in the \textit{open-world} setting, where a robot learns to manipulate novel objects from a single video demonstration. We introduce \ourmethod{}, an algorithm that tackles the problem by extracting an object-centric manipulation plan from a single RGB or RGB-D video and deriving a policy that conditions on the extracted plan. Our method enables the robot to learn from videos captured by daily mobile devices and to generalize the policies to deployment environments with varying visual backgrounds, camera angles, spatial layouts, and novel object instances. We systematically evaluate our method on both short-horizon and long-horizon tasks, using RGB-D and RGB-only demonstration videos. Across varied tasks and demonstration types (RGB-D / RGB), we observe an average success rate of 74.4\%, demonstrating the efficacy of ORION in learning from a single human video in the open world. Additional materials can be found on the \href{https://ut-austin-rpl.github.io/ORION-release}{project website}.
\end{abstract}

\newcommand\blfootnote[1]{%
  \begin{NoHyper}%
  \renewcommand\thefootnote{}\footnote{#1}%
  \addtocounter{footnote}{-1}%
  \end{NoHyper}%
}

\blfootnote{*Equal Contribution.}

\section{Introduction}
\label{sec:intro}
\input{01_intro}

\section{Problem Formulation}
\label{sec:problem}
\input{03_problem}

\section{Method}
\label{sec:method}
\input{04_method}

\section{Experiments}
\label{sec:experiments}
\input{05_experiments}

\section{Related Work}
\label{sec:related-works}
\input{02_related}

\section{Conclusions}
\label{sec:conclusions}
\input{06_conclusions}

\section*{Acknowledgments} We would like to thank Rutav Shah, Jiayuan Mao, Fangchen Liu, Bowen Wen, and Stan Birchfield for helpful discussions. This work has taken place in the Robot Perception and Learning Group (RPL) and Learning Agents Research Group (LARG) at UT Austin. RPL research has been partially supported by the National Science Foundation (FRR2145283, EFRI-2318065) and the Office of Naval Research (N00014-22-1-2204). LARG research is supported in part by the National Science Foundation (FAIN-2019844, NRT-2125858), the Office of
Naval Research (N00014-24-1-2550), Army Research Office
(W911NF-17-2-0181, W911NF-23-2-0004, W911NF-25-1-0065), DARPA
(Cooperative Agreement HR00112520004 on Ad Hoc Teamwork), Lockheed
Martin, and Good Systems, a research grand challenge at the University
of Texas at Austin.  The views and conclusions contained in this
document are those of the authors alone.  Peter Stone serves as the
Chief Scientist of Sony AI and receives financial compensation for
that role.  The terms of this arrangement have been reviewed and
approved by the University of Texas at Austin in accordance with its
policy on objectivity in research.

\newpage

\bibliography{references}

\appendix
\input{supp}

\end{document}

%% file: 01_intro.tex
A critical step toward building robot autonomy is developing sensorimotor skills for perceiving and interacting with unstructured environments. Conventional methods for acquiring skills necessitate manual engineering and/or costly data collection~\cite{dalal2021accelerating, mandlekar2020learning, nasiriany2022augmenting, zhang2023noir, zhu2022viola}. A promising alternative is teaching robots through human videos of manipulation behaviors situated in everyday scenarios. These methods have great potential to tap into the readily available source of Internet videos that encompass a wide distribution of human activities, paving the way for scaling up skill learning. 

Prior work on learning from human videos has focused on pre-training representations and value functions~\cite{chen2021learning, nair2022r3m, ma2022vip, wang2023mimicplay, xiong2021learning}. However, they do not explicitly capture object interactions and states in 3D space where robot motions are defined. Consequently, they require separate teleoperation data for each set of objects in each location and even for each possible change in visual backgrounds, \textit{e.g.}, the scene background or lighting conditions~\cite{zhu2023groot}. In contrast, our goal is for a robot to learn to perform a task robustly in the ``open world", \textit{i.e.}, under varying visual and spatial conditions from a single human video, without prior knowledge of the object models or the behaviors shown. Since our policy construction process uses actionless videos that are equivalent to state-only demonstrations in the problem of ``imitation from observation"\cite{torabi2021imitation}, we refer to our problem setting as \textit{open-world imitation from observation}.

Developing such a method in this setting becomes possible due to the recent advances in vision foundation models~\cite{kirillov2023segment, oquab2023dinov2}. These models, pre-trained on Internet-scale visual data, excel at understanding open-vocabulary visual concepts and enable robots to recognize and localize objects in natural videos without known object categories or access to physical states. This work marks a major step toward achieving our vision of open-world imitation from observation, where a robot learns to interact with objects given \textit{a single video} while generalizing to environments with different visual backgrounds and unseen spatial configurations during deployment. In this work, we consider using RGB or RGB-D video demonstrations where a person manipulates a small set of task-relevant objects, recorded with a stationary camera. These videos are actionless or state-only, as they do not come with any ground-truth action labels for the robot.

\begin{figure}
    \centering    \includegraphics[width=\linewidth]{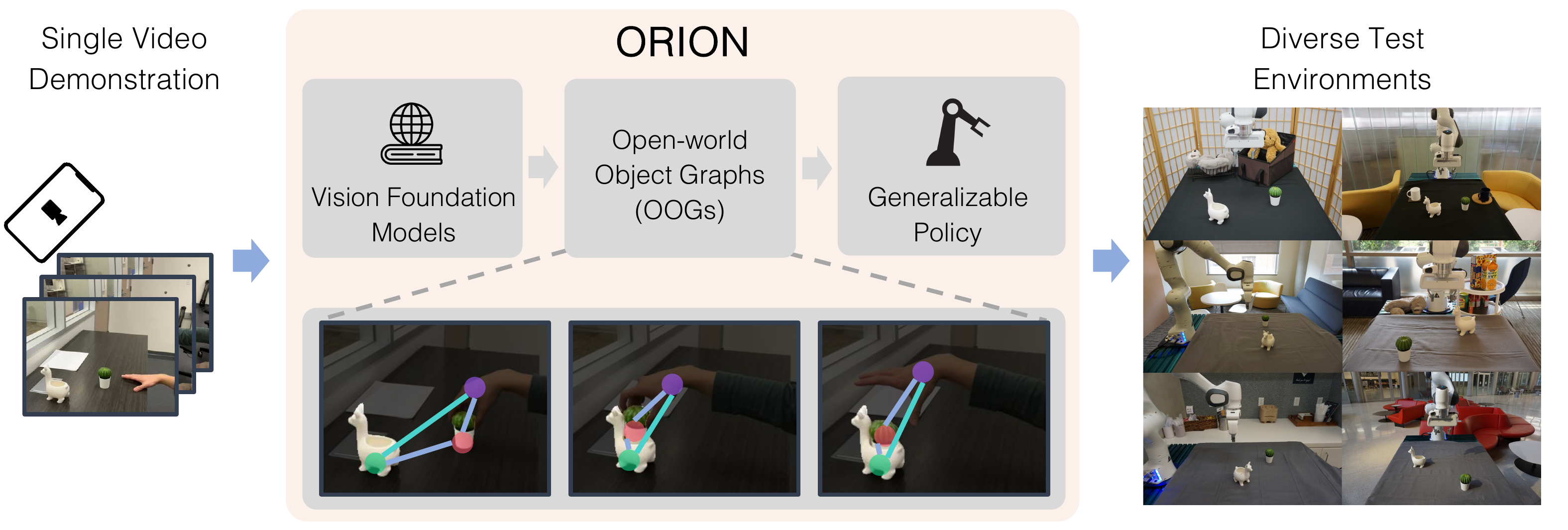}
    \caption{\textbf{Overview.} We introduce \ourmethod{} for tackling the problem of learning manipulation behaviors from single human video demonstrations. \ourmethod{} first extracts a sequence of Open-World Object Graphs (OOGs), where each OOG models a keyframe state with task-relevant objects and hand information. Then, \ourmethod{} leverages the OOG sequence to construct a manipulation policy that generalizes across varied initial conditions, specifically in four aspects: visual background, camera shifts, spatial layouts, and novel instances from the same object categories. }
    \label{fig:pull-figure}
    \vspace{-2mm}
\end{figure}

We introduce our method \ourmethod{}, short for \textbf{O}pen-wo\textbf{R}ld video \textbf{I}mitati\textbf{ON}. Figure~\ref{fig:pull-figure} visualizes a high-level overview of \ourmethod{}. The core innovation lies in creating an object-centric spatiotemporal abstraction that effectively bridges the observational gap between human demonstration and robot execution. The design of \ourmethod{} stems from our insight that manipulation tasks center around object interaction, and task completion depends on whether specific intermediate states, so-called \textit{subgoals}, are reached. To capture the object-centric information in the video, we design a graph-based, object-centric representation, called Open-world Object Graphs (OOGs), to model the states of task-relevant objects and their relationships. An OOG has a two-level hierarchy. The high level consists of object nodes and a grasp node, where object nodes identify and localize the relevant objects by leveraging outputs from vision foundation models, while the grasp node encodes the interaction information between the hand and objects, such as where to grasp. The low level consists of point nodes, which correspond to object keypoints, and the node features detail the motions of object keypoints in the 3D space.

\ourmethod{} extracts a manipulation plan from the video as a sequence of OOGs and uses the plan to construct a generalizable policy. Experiments indicate that \ourmethod{} constructs manipulation policies that are robust to conditions vastly different from those in the video.  Using only an iPhone or iPad recording — or even synthetic videos generated by models such as Veo 2 \cite{veo2} — of a task being performed in everyday environments (\textit{e.g.}, an office or a kitchen), the resulting policies succeed in workspaces with drastically different visual backgrounds, camera angles, and spatial arrangements, and even generalize to manipulating unseen object instances of the same categories.

In summary, our contribution is threefold. 1) We formulate the problem of learning vision-based robot manipulation from a single human video in the open-world setting; 2) We introduce Open-world Object Graphs (OOGs), a graph-based, object-centric representation that models the states and relations of task-relevant objects; and 3) We present \ourmethod{}, an algorithm that 
uses a single video to construct a manipulation policy, which generalizes to conditions that differ in four key ways: visual backgrounds, camera perspectives, spatial configurations, and new object instances.

%% file: 03_problem.tex
In this paper, we consider a vision-based, tabletop manipulation task, formulated as a finite-horizon Markov Decision Process (MDP) described by a tuple $\langle \mathcal{S}, \mathcal{A}, \mathcal{P}, H, R, \mu \rangle$, where $\mathcal{S}$ is the state space of raw sensory data and robot proprioception, $\mathcal{A}$ is the action space of low-level robot commands, $\mathcal{P}: \mathcal{S} \times \mathcal{A} \mapsto \mathcal{S}$ is the transition dynamics, $H$ is the maximal task horizon, $R$ is the sparse reward function, and $\mu$ is the initial state distributions of a task. We focus on the case where task reward functions are defined based on the object relations (contact and non-contact relations) between a small set of \emph{task-relevant objects}. For example, a contact relation change occurs when a mug is placed on top of a coaster. A non-contact relation change occurs when a cup is angled to pour liquid into a bowl. A reward function returns 1 if all object relations of a task are satisfied and 0 otherwise. The primary objective of solving a manipulation task is to find a visuomotor policy $\pi$ that maximizes the expected task success rate from a wide range of initial environment configurations characterized by $\mu$, where the states vary across the following four dimensions: 1) changing visual backgrounds, 2) different camera angles, 3) different object instances from the same categories, and 4) varied spatial layouts of the task-relevant objects. 

We assume a robot does not have direct access to the ground-truth task reward or the physical states of task-relevant objects. In addition, we assume a single \textit{actionless} video~\cite{ko2023learning, wen2023any}. $V$ is provided as a \textit{state-only} demonstration consisting of an RGB or RGB-D video stream of a person manipulating task-relevant objects, captured as an arbitrarily long sequence of images using a stationary camera. In addition, $V$ involves a manipulation sequence where the object relations among task-relevant objects or hands change (\textit{e.g.}, an object is gripped, then placed on top of another object, or an object is tilted into a pouring position over another object). In this scenario, however, the robot is not pre-programmed to have access to ground-truth categories and locations of the task-relevant objects in $V$. We refer to this challenging setting as ``open-world''~\cite{joseph2021towards}, as the robot must imitate from $V$ while not pre-programmed or trained to interact with the objects in $V$. To allow a robot to operate in this "open-world'' setting, we assume access to common sense knowledge through large models pre-trained on internet-scale data, \textit{i.e.}, foundation models. 

For evaluation, we adopt the following procedure. Given a single video $V$ that matches the object relations of a task instance drawn from $\mu$, the model's performance is measured by the average rewards received when evaluating new task instances drawn from the same $\mu$. If the manipulation in $V$ does not result in object relations that satisfy the task's requirements (\textit{e.g.}, the task specifies pouring juice into a cup, but the video pours juice on the floor), it is not considered a valid demonstration and thus should not be used for evaluation.

%% file: 04_method.tex
In this section, we describe our method~\ourmethod{} (\textbf{O}pen-wo\textbf{R}ld video \textbf{I}mitati\textbf{ON}). \ourmethod{} is an algorithm that allows a robot to mimic human execution of a task, given a single demonstration video $V$. To effectively construct a policy $\pi$ from $V$, \ourmethod{} employs a learning objective based on an object-centric prior. The goal is to create a policy $\pi$ that directs the robot to move objects along trajectories that mimic the directional and curvature patterns observed in $V$, relative to the objects' initial and final positions. This objective is based on the observation that objects are likely to achieve target configurations by moving along trajectories similar to those in $V$. Key to \ourmethod{} is generating a manipulation plan from $V$, which serves as the spatiotemporal abstraction of the video that guides the robot to perform a task. A plan is an arbitrarily long sequence of object-centric keyframes, where each keyframe specifies an initial or a subgoal state captured in $V$. We first introduce our formulation of the object-centric representation of a state, Open-world Object Graph (OOG), used in \ourmethod{}, and then describe the algorithm that constructs a robot policy given a human video.\loosepar{}

\begin{figure*}[t]
    \centering
    \includegraphics[width=1.0\linewidth, trim=0cm 0cm 0cm 0cm,clip]{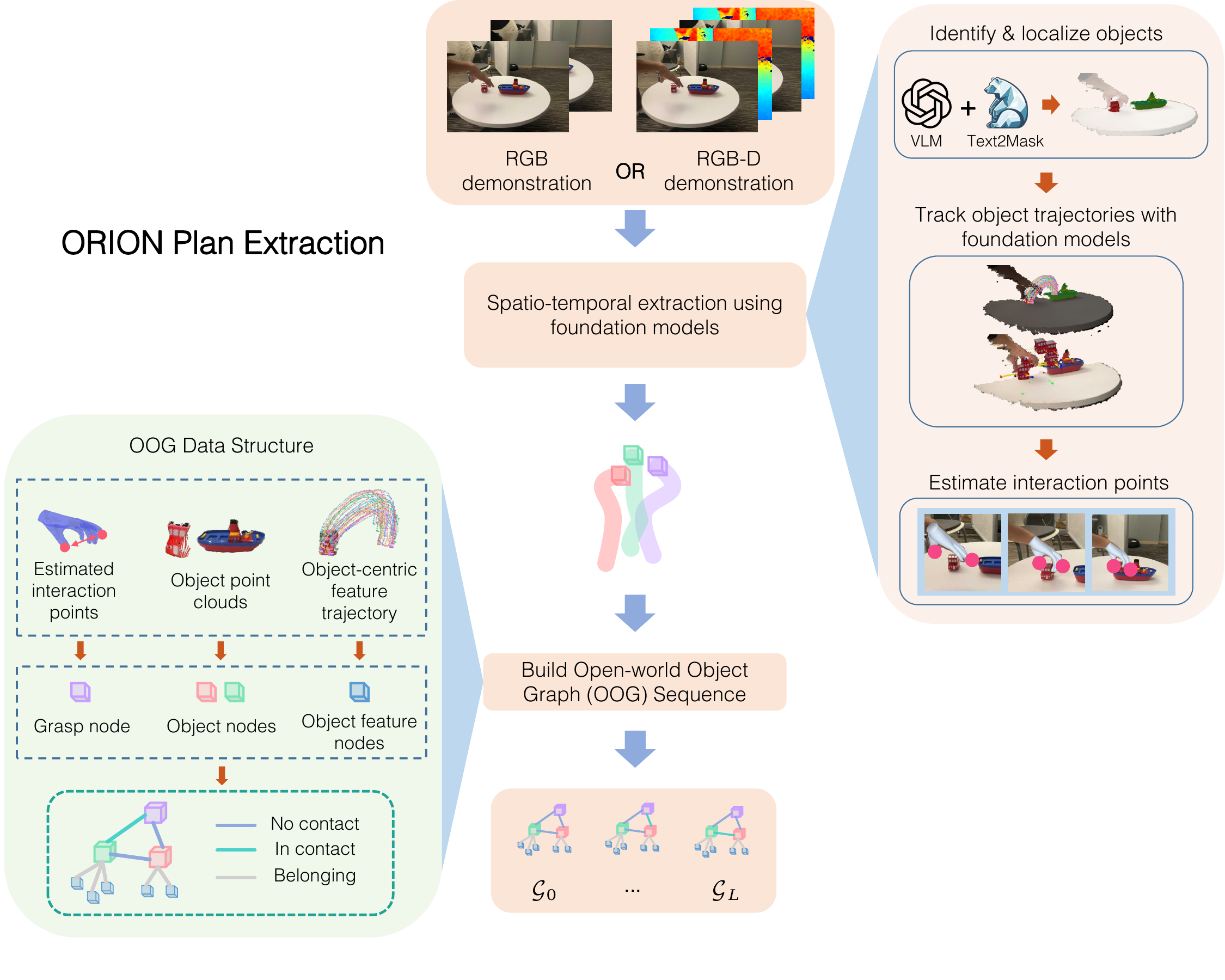}
    \vspace{-1mm}
    \caption{\textbf{Overview of plan generation in \ourmethod{}.} \ourmethod{} generates a manipulation plan from a given video $V$ in order for subsequent policies to synthesize actions. \ourmethod{} first tracks objects and keypoints across the video frames. Then, keyframes are identified based on the velocity statistics of the keypoint trajectories. Finally, \ourmethod{} generates an  Open-world Object Graph (OOG) for every keyframe, resulting in a sequence of OOGs that serves as the spatiotemporal abstraction of the video.
    The figure is best viewed in color. 
    }
    \label{fig:model-part1}
    \vspace{-2mm}
\end{figure*}

\subsection{Open-world Object Graph}
At the core of our approach is a graph-based, object-centric representation, Open-world Object Graphs (OOGs). OOGs use open-world vision models to model visual scenes with task-relevant objects and manipulators (e.g., a human hand or gripper) such that they naturally exclude the distracting factors in visual data and localize the task-relevant objects regardless of their spatial locations (see Section~\ref{sec:plan-generation}).

We denote an OOG as $\G$, which consists of object, gripper, and point nodes. At the high level, each object node corresponds to a task-relevant object described by the output of open-world vision models. Every object node comes with node features,  consisting of colored 3D point clouds derived from RGB and depth observations. These node features indicate both what and where objects are and also represent their geometry information. For RGB-only videos, an object node additionally includes an estimated object mesh, along with depth estimated from monocular depth estimation models. Next, to inform the robot where to interact with objects (\textit{e.g.}, where to grasp), we introduce the specialized ``grasp node'', which stores the interaction cues such as contact points and the grip status (open or closed) that can be directly mapped to the robot end-effector during execution. At the low level, each point node corresponds to a keypoint that belongs to a task-relevant object.  Every point node comes with an object-centric feature trajectory. This trajectory contains a list of features describing the object's state throughout the video, such as a list of 3D points or 6-DOF poses. The information in this trajectory explicitly models how an object should be moved during a manipulation task. In the rest of the paper, by motion features of a point node in $G_l$, we mean the object-centric feature trajectory between keyframe $l$ and $l+1$. 

In an OOG, all the object nodes and the hand node are fully connected, reflecting real-world spatial relationships. Additionally, the edges are augmented with a binary attribute, indicating whether two objects or an object and the hand are in contact. This attribute allows our developed algorithm to check the set of contact relations that are satisfied, retrieving the matched OOG from the generated plan (see Section~\ref{sec:plan-generation}). As for the low-level point nodes, they are connected to their respective object node, indicating a belonging relationship. In the rest of the paper, we denote node entities from human videos with a superscript $\human$, and denote the ones from the robot rollout with a superscript $\robot$. Table~\ref{tab:oog_data_structure} in the appendix also summarizes the variables needed to define an OOG.\loosepar{}

\begin{algorithm}[t]
\caption{Manipulation Plan Generation from Video $V$}
\label{alg:plan-generation}
\begin{algorithmic}[1]
\Require RGB frames $r_{1...N}$ from demonstration video $V$ 
\Optional Depth frames $d_{1..N}$ from demonstration video $V$
\Ensure OOG sequence $\{\G_0,\dots,\G_L\}$
\Statex \textbf{// 1. Track task‐relevant objects}
\State $\mathit{obj\_list} \leftarrow$ \Call{VLM}{$r$}
\State $\mathit{masks}_1 \leftarrow$ \Call{GroundedSAM}{$r_1$, $\mathit{obj\_list}$}
\State $\mathit{masks}_{2..N} \leftarrow$ \Call{Cutie.track}{$\mathit{masks_1}$, $r_{2...N}$}
\If{$d \neq \varnothing$} \Comment{Depth available}
    \State $\mathit{keypoints} \leftarrow$ \Call{sample\_points}{$\textit{masks}_1$}
    \State $\textit{trajectories} \leftarrow$ \Call{CoTracker.track}{$\textit{keypoints}$, $r$} \Comment{Track object keypoints in $V$}
\Else \Comment{
    No depth available
}
    \State $d \leftarrow$ \Call{estimate\_depth}{$r$}
    \State $\mathit{meshes} \leftarrow$  \Call{scale\_from\_depth}{\Call{InstantMesh}{$\mathit{mask}$, $r$}, $d$} \Comment{Generate object meshes}
  \State $\mathit{trajectories} \leftarrow$ \Call{FoundationPose}{$r$, $d$, $\mathit{masks}$, $\mathit{meshes}$} \Comment{Track object poses in $V$}
\EndIf

\Statex \textbf{// 2. Discover keyframes}
  \State $\textit{keyframes}$ $\leftarrow$ \Call{changepoint\_detect}{$\textit{trajectories}$}

\Statex \textbf{// 3. Generate OOGs}
\State $T_{table} = $ \Call{estimate\_plane}{$d_1$}
\For{each keyframe $l$ in keyframes}
    \State $\G_l \leftarrow$ \Call{empty\_graph}{\ }
    \State \Call{add\_object\_nodes}{$\G_l$, $\textit{masks}_l$, $d_l$, $\mathit{meshes}$ \textbf{if} $meshes \neq \varnothing$}
    \State \Call{add\_point\_nodes}{$\G_l$, $\mathit{trajectories}_l$}
    \State \Call{add\_grasp\_node}{$\G_l$, $d_l$, $\mathit{meshes}$ \textbf{if} $meshes \neq \varnothing$}
    \State $G_l \leftarrow $ \Call{transform}{$T, G_l$} \Comment{Align OOG features with plane}
\EndFor

\State \Return $\{\G_l\}_{l=0}^L$
\end{algorithmic}
\end{algorithm}

\subsection{Manipulation Plan Generation From $V$}
\label{sec:plan-generation}

We describe the first part of \ourmethod{} (see Figure~\ref{fig:model-part1} and Algorithm~\ref{alg:plan-generation}), where our method automatically annotates the video and generates a manipulation plan from a given human video, $V$. In this paper, a manipulation plan is a spatio-temporal abstraction of $V$ that centers around the object states and their motions over time, demonstrating how a task should be completed. Our core insight is that we can cost-effectively model a task with object locations at some keyframe states where the set of satisfied contact relations is changed, and abstract a majority of intermediate states into object motion trajectories (3D translations or 6-DOF pose trajectories). Concretely, a plan is represented as a sequence of OOGs, $\{ \G_l \}_{l=0}^{L}$, which corresponds to $L + 1$ keyframes in $V$, with $\G_0$ representing the initial state. Note that \ourmethod{} processes demonstration videos differently depending on whether depth information is available. We refer to these as the \textit{RGB-D} case (with RGB and depth) and the \textit{RGB} case (with RGB only).

\myparagraph{Tracking task-relevant objects.} \ourmethod{} first localizes task-relevant objects in the video $V$ (Lines 1-3). To begin, \ourmethod{} samples frames from $V$ and passes these frames to a VLM such as GPT-4o.  The VLM is then prompted to return a list of strings describing each task-relevant object, while also indicating whether an object is being manipulated (details in Appendix \ref{supp:prompt}). This list is then used as an input to an open-world vision model, Grounded-SAM~\cite{liu2023grounding}, to annotate video frames with segmentation masks of the task-relevant objects. In practice, open-world vision models are computationally demanding, so we reduce the computation by exploiting object permanence to track the objects (Lines 4-10). Specifically, \ourmethod{} annotates the first video frame with Grounded-SAM, and then propagates the segmentation to the rest of the video using a Video Object Segmentation Model, Cutie~\cite{cheng2023putting}. In the RGB case, \ourmethod{} instead tracks an object's 6-DOF trajectory with a modified version of FoundationPose \cite{wen2024foundationpose}. The input to FoundationPose includes estimated depth, per-frame object masks, and an estimated object mesh, which is generated by InstantMesh \cite{xu2024instantmesh} (implementation details in Appendix \ref{supp:implementation}).\loosepar{}

\myparagraph{Discovering keyframes.} After annotating the locations of task-relevant objects, \ourmethod{} tracks their motions across the video to discover $K$ keyframes based on the velocity statistics of object motions (Line 12). This design is based on the observation that changes in object contact relations due to manipulation are often accompanied by sudden changes in object motions  (\textit{e.g.}, transitioning from free space motion to grasping an object). However, keeping full track of object point motion using techniques like optical flow estimation often results in unsatisfactory tracking quality, largely due to occlusions during manipulation. \ourmethod{} uses a track-any-point (TAP) model, namely CoTracker~\cite{karaev2023cotracker}, to track a subset of points in a long-term video with explicit occlusion modeling, which has been successfully applied to track object motions in robot manipulation~\cite{vecerik2023robotap, wen2022you}. Specifically, \ourmethod{} first samples keypoints within the object segmentation of the first frame and track the trajectories across the video. The changes in velocity statistics are straightforward to detect based on the TAP trajectories, where we discover the keyframes using a standard unsupervised changepoint detection algorithm~\cite{killick2012optimal}. In the RGB case, we find that setting $K=2$, where the first keyframe is the first video frame and the last keyframe is the last video frame, still results in an OOG sequence that sufficiently describes the object motions in $V$.


\myparagraph{Generating OOGs from $V$. } Once \ourmethod{} discovers the keyframes, it generates an OOG at each keyframe to model the state of task-relevant objects and the human hand in $V$. The creation of OOG nodes can reuse the results from the annotation process. For object nodes, the point clouds for node features are obtained by back-projecting the object segmentation with depth data (Line 16). In the RGB-D case, each point node corresponds to sampled keypoints. Their motion features, 3D trajectories, are back-projected from the TAP trajectories using depth data (Line 17). In the RGB case, we have a single point node containing a 6-DOF object trajectory generated by FoundationPose.

Additionally, grip information is required to specify the interaction points with task-relevant objects and the grip status to be mapped to the robot gripper (Line 18). In the RGB-D case, we use a hand-reconstruction model, HaMeR~\cite{pavlakos2023reconstructing}, which gives a reconstructed hand mesh that pinpoints the hand locations at each keyframe. The distances between the fingertips of the mesh help determine the grip status, \textit{i.e.}, whether it is open or closed. In the RGB case, since many in-the-wild videos do not show a human hand or significantly occlude the hand, \ourmethod{} uses grasp heuristics extracted from the estimated object mesh or point cloud, combined with the estimated 6-DOF pose.

With all the node information, \ourmethod{} establishes the edge connections between nodes in OOGs, representing contact relations. Since all object and hand locations are computed in the camera frame while the camera extrinsic of $V$ is unknown, there is ambiguity when deciding the spatial relations between objects. We exploit the assumption of tabletop manipulation, where a plane is always present with its normal direction aligned with the z-axis of the world coordinate system (Lines 13, 19). \ourmethod{} estimates the transformation matrix of the table plane and transforms all the point cloud features in OOGs to align with the x-y plane of the world coordinate (Full details appear in Appendix~\ref{supp:implementation}). Then, the contact relations in each state can be determined based on the spatial relations and the computed distances between point clouds. The relations allow \ourmethod{} to match the test-time observations with a keyframe state from the plan and subsequently decide which object to manipulate (see Section~\ref{sec:plan-execution}). In the end, \ourmethod{} generates a complete OOG for each discovered keyframe. 

\begin{figure*}[t]
    \centering
    \includegraphics[width=1.0\linewidth, trim=0cm 0cm 0cm 0cm,clip]{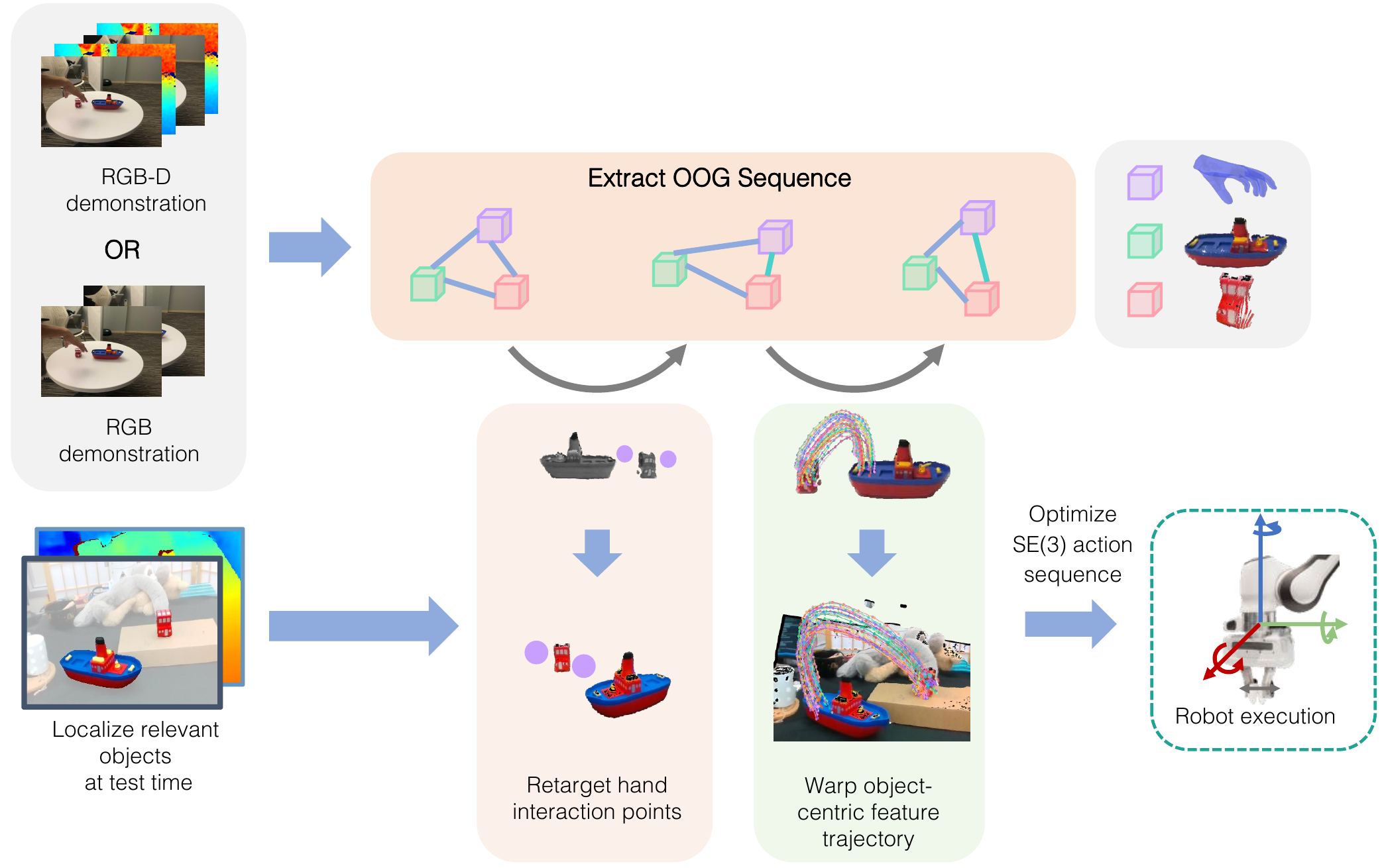}
    \vspace{-2mm}
    \caption{\textbf{Overview of the \ourmethod{} Policy}. \ourmethod{} first localizes task-relevant objects at test time and retrieves the matched OOG from the generated manipulation plan. Then, \ourmethod{} uses the retrieved OOGs to predict the object motions by warping the object-centric feature trajectory from the video to match the test-time observation. The predicted trajectories are then used to optimize the SE(3) action sequence of the robot end effector, which is subsequently used to command the robot.}
    \label{fig:model-part2}
    \vspace{-3mm}
\end{figure*}

\begin{algorithm}[t]
\caption{Robot Policy to Synthesize Actions}
\label{alg:plan-execution}
\begin{algorithmic}[1]
\Require Manipulation plan $\G = \{\G_0,\dots,\G_L\}$, Demonstration video $V$
\Ensure Sequence of robot end‑effector SE(3) actions, executed on robot
\While{not \textit{task\_completed} and not \textit{task\_failed}}
  \Statex \textbf{// 1. Retrieve matching OOGs from plan}
  \State $r^{Ro}$, $d^{Ro} \leftarrow$  \Call{get\_robot\_obs}{\ }
  \State $\G^{\robot} \leftarrow$ \Call{generate\_OOG}{$r^{Ro}, d^{Ro}$} \Comment{Same vision pipeline as in ~\ref{sec:plan-generation}}
  \State $(\G_{l}, \G_{l+1}) \leftarrow $ \Call{find\_matching\_OOG}{$\G^{Ro}, \G$}
  \State $\hat{o}_{target}$, $\hat{o}_{ref}$, $\tau^V_{target} \leftarrow$ \Call{extract\_objs\_demo}{$\G_l$}
  \State $o_{target}$, ${o}_{ref} \leftarrow$ \Call{extract\_objs\_test}{$\hat{o}_{target}$, $\hat{o}_{ref}$, $\G_l$, $\G_{l+1}$, $\G^{Ro}$}

  \Statex \textbf{// 2. Predict object motions}
  \State $s_{new}, e_{new} = \Call{find\_rollout\_start\_end}{\hat{o}_{target}, {o}_{target}, \hat{o}_{ref}, {o}_{ref}}$
  \State $\tau^{Ro}_{target} \leftarrow $ \Call{warp\_trajectories}{$\tau^V_{target}, s_{new}, e_{new}$} \Comment{Details in Algorithm \ref{alg:trans-warp} and \ref{alg:rot-warp}}

  \Statex \textbf{// 3. Optimize robot actions}
  \State $\{T_i\}_{i=0}^{t_{l+1}-t_l}$ $\leftarrow$ $\argmin\limits_{T_0, T_1, \dots,  T_{t_{l+1} - t_{l}}} \sum_{i=0}^{t_{l+1} - t_{l}}\norm{\tau^{\robot}_{\target}({i+1}) - T_{i}\tau^{\robot}_{\target}({i})}_2^2
$

  \State $\mathit{joint\_seq} \leftarrow $ \Call{IK}{$\{T_i\}$}  
  \State $\mathit{joint\_seq} \leftarrow $ \Call{add\_grip\_info}{$\mathit{joint\_seq}$, $\G_l.vh$}

  \State \Call{robot\_execution}{joint\_seq}

\EndWhile
\end{algorithmic}
\end{algorithm}

\subsection{Robot Policy To Synthesize Actions}
\label{sec:plan-execution}
Given a manipulation plan, \ourmethod{} constructs a policy that synthesizes actions, detailed in Figure~\ref{fig:model-part2} and Algorithm ~\ref{alg:plan-execution}. The manipulation policy is derived based on the aforementioned learning objective to achieve object motion similarities. The action synthesis comprises three major steps: identify a keyframe from the plan that matches the current observation, predict object motions, and use the predictions to optimize the robot actions for the robot controller to execute. The policy repeats these three steps until a task is completed or fails. Task completion conditions are detailed in Appendix~\ref{supp:success}.

\myparagraph{Retrieving OOGs from the plan.} \ourmethod{} first identifies the keyframe and retrieves OOGs to help decide what next actions to take (Lines 2-4). At test-time, \ourmethod{} localizes objects in the new observations and estimates contact relations using the same vision pipeline as described in Section~\ref{sec:plan-generation}. Then, \ourmethod{} retrieves the OOG that has the same set of relations as the current state, allowing us to identify a pair $(\G_{l}, \G_{l+1})$, where $\G_{l}$ is the retrieved graph and $\G_{l+1}$ the graph of the next keyframe.  This pair of graphs provides sufficient information to decide which object to manipulate next, termed the \textit{target object}, and we denote its point cloud/pose at keyframe $l$ as $\hat{o}_{\target}$, and its object-centric feature trajectories as $\tau^{\human}_{\target}$ (Lines 5-6). A target object is the one in motion due to manipulation between two keyframes. In the RGB-D case, the target object is identified by computing the average velocity per-object using motion features in $G_{l}$. At the same time, another object, called the \emph{reference object}, is involved in changing contact state relations from $\G_{l}$ to $\G_{l+1}$ and serves as a spatial reference for the target object's movement. In the RGB case, \ourmethod{} uses the outputs of the classification done by the VLM to determine which objects are target and reference objects. In both cases, \ourmethod{} use the point cloud of the reference object at the \textit{next keyframe} $l+1$, as the reference object might have location changes due to object interactions, and using the updated information from the next keyframe gives us an accurate prediction of the trajectories. Once the target and reference objects are determined, we can localize the corresponding objects in the test-time observations, where their point clouds/poses are denoted as  $o_{\target}$ and $o_{\reference}$, respectively.

\myparagraph{Predicting object motions.} Given the target and reference objects from keyframes $l$ and $l+1$, \ourmethod{} predicts the motion of the target object in the current state by warping the object-centric feature trajectories estimated from $V$ (Lines 7-8). To warp the trajectories, \ourmethod{} first identifies the initial and goal locations of keypoints in the new configuration by leveraging information given by the OOG pair. In the RGB-D case, \ourmethod{} uses global registration of point clouds~\cite{choi2015robust} to align $\hat{o}_{\target}$ with $o_{\target}$ and $\hat{o}_{\reference}$ with $o_{\reference}$, giving two transformations to compute the new starting and goal positions of target object keypoints conditioned on where the reference object is. In the RGB case, since we have the 6-DOF poses of $\hat{o}_{\target}$,  $o_{\target}$, $\hat{o}_{\reference}$, and $o_{\reference}$, \ourmethod{} directly finds the relative transformations required to align the respective pairs.

Now, given new starting and goal positions of the target objects, \ourmethod{} must warp the original object-centric feature trajectory to fit these new start and end conditions. In the RGB-D case, \ourmethod{} is warping a 3D point trajectory, which we call \textit{translation warping} (full details in Algorithm \ref{alg:trans-warp}). \ourmethod{} first normalizes $\tau^{\human}_{\target}$ with its starting and goal locations, obtaining $\hat{\tau}_{\target}$. $\hat{\tau}_{\target}$ only contains the directional and curvature patterns that are independent of the absolute location of the initial and the goal keypoints. Then, $\hat{\tau}_{\target}$ is scaled back to the workspace coordinate frame using the new starting and goal locations, resulting in new keypoint trajectories of the target object $\tau^{\robot}_{\target}$. 

In the RGB case, since we have a 6-DOF pose trajectory, \ourmethod{} also needs to do \textit{rotation warping} in addition to translation warping (full details in Algorithm \ref{alg:rot-warp}).  In rotation warping, \ourmethod{}\  first computes the alignment rotation $A$ that maps the demonstration's reference‑to‑target object direction to the test-time reference-to-target object direction, and left-multiplies $A$ on every rotation in  $\tau^{\human}_{\target}$. \ourmethod{} then normalizes the resulting sequence about its first frame—so it begins at the identity rotation—yielding a ``relative rotation'' trajectory.

Next, \ourmethod{} scales this normalized sequence back to the workspace by defining two boundary rotations: one that maps the normalized start to the real initial pose, and one that maps the normalized end to the real goal pose. \ourmethod{} uses SLERP between these two rotations to get a smooth transition $\{I_i\}$, and post‑multiplies the normalized rotations by $I_i$, at each timestep.  This set of operations guarantees the warped rotation begins and ends exactly at the real‑scene boundary conditions, while still containing the same rotation trajectory ``shape'' from the original $\tau^{\human}_{\target}$. 

\myparagraph{Optimizing robot actions.} Once we obtain $\tau^{\robot}_{\target}$, we optimize for a sequence of SE(3) transformations that guide the robot end-effector to move. The SE(3) transformations are optimized to align the keypoint locations from previous frames to the next frames along the predicted trajectories:
\begin{equation}
    \argmin_{T_0, T_1, \dots,  T_{t_{l+1} - t_{l}}} \sum_{i=0}^{t_{l+1} - t_{l}}\norm{\tau^{\robot}_{\target}({i+1}) - T_{i}\tau^{\robot}_{\target}({i})}_2^2
    \label{label:optim}
\end{equation}
where $\tau^{\robot}_{\target}(i)$ $(0 \leq i \leq t_{l+1} - t_{l})$ represents the keypoint locations at timestep $i$ along the trajectory (Line 9) and $t_l$ represents the time index for the keyframe $l$. This optimization process naturally allows generalizations over spatial variations, as the action sequence always conditions on a new location instead of overfitting to absolute locations. In the RGB case, however, because \ourmethod{} already has the warped 6-DOF pose trajectory of the object, we no longer need to solve Equation \ref{label:optim}  and can find $T_i$ in the 6-DOF object pose trajectory.

To further specify where the gripper should interact with the object and whether it should be open or closed, \ourmethod{} augments the resulting SE(3) sequence with the interaction information stored in the grip node $\HandNode$ (Line 11). We implement a combination of inverse kinematics (IK) and joint impedance control to achieve precise and compliant execution (Lines 10, 12). 

The resulting \ourmethod{} policy is robust to visual variations due to the use of open-world vision models. It also generalizes to different spatial locations due to our choice of representing object locations in object-centric frames and the optimization process that is not constrained to specific positions. \loosepar{}

%% file: 05_experiments.tex
In this section, we report on experiments to answer the following questions regarding the effectiveness of \ourmethod{} and the important design choices: 

\begin{enumerate}
    \item Is \ourmethod{} effective in constructing manipulation policies from single human videos in the open-world setting?
    \item To what extent do the object-centric abstractions improve the policy performance?
    \item How critical is it to model the object motions with keypoints and the TAP formulation?
    \item How consistent is \ourmethod{}'s performance on videos taken in various conditions? 
    \item Can \ourmethod{} scale to multi-step, long-horizon manipulation tasks?
    \item  Can \ourmethod{} effectively construct manipulation policies from RGB-only demonstration videos in the absence of depth information? 
\end{enumerate}        

\subsection{Experiment Setup}

\begin{figure}
    \centering
    \includegraphics[width=\linewidth]{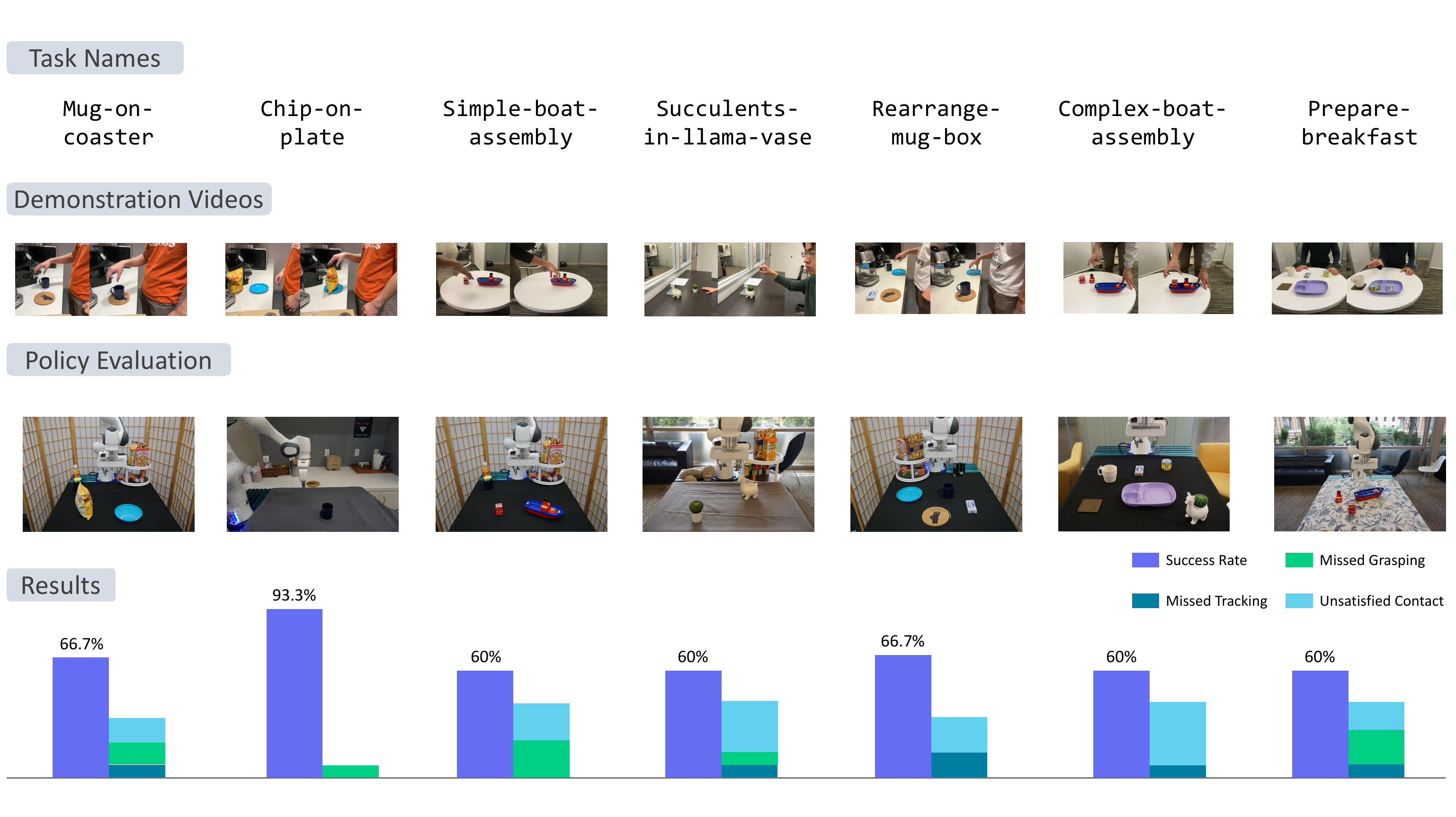}
    \caption{The upper part of the figure illustrates the following items: the initial and final frames of human videos for every task, the list of word descriptions provided along with the video, and the example images of initial states for policy evaluation. The lower part of the figure shows the overall evaluation of \ourmethod{} over all seven tasks, including the success rates and the quantification of failed trials, separated by failure mode.}
    \vspace{2mm}
    \label{fig:tasks}
\end{figure}

\myparagraph{Task descriptions.} We design the following seven tasks to evaluate the policy performance: 

\begin{enumerate}
    \item \mugcoaster{}: placing a mug on the coaster.
    \item \simpleboat{}: putting a small red block on a toy boat.
    \item \chip{}: placing a bag of chips on the plate.
    \item \llama{}: inserting succulents into the llama vase.
    \item \juice{}: pouring juice into a container.
    \item \peasplate{}: putting a can of peas on a plate.
    \item \cheeseplate{}: putting a block of cheese on a plate.
    \item \rearrange{}: placing a mug on a coaster and placing a cream cheese box on a plate consecutively.
    \item \complexboat{}: placing both a small red block and a chimney-like part on top of a boat.
    \item \breakfast{}: placing a mug on a coaster and putting a food box and can on the plate. 
\end{enumerate}

The first seven are ``short-horizon'' tasks, and the last three are ``long-horizon'' tasks.  In the context of this paper, ``short-horizon'' refers to tasks that only require one contact relation between two objects, while ``long-horizon'' refers to those that require more than one contact relation. Finally, we note that \mugcoaster{} and \llama{} have both RGB and RGB-D demonstration videos, while \juice{}, \peasplate{}, and \cheeseplate{} only have RGB demonstration videos.

Detailed success conditions of all tasks are described in Appendix~\ref{supp:success}.

\myparagraph{Experimental setup.} We design experiments to fully test the efficacy of our method by providing the robot with videos captured in everyday scenarios, which naturally encompass visual backgrounds and camera setups that are different from those of the robot. Specifically, we record an RGB-D video of a person performing each of the seven tasks in everyday scenarios, such as an office or a kitchen. We use an iPad for recording, which comes with a TrueDepth Camera, and we fix it on a camera stand. The videos can be found on the project website. During test time, the robot receives visual data through a single RGB-D camera, Intel RealSense D435, and performs manipulation in its workstation to evaluate policies. We use the 7-DoF Franka Emika Panda robot for all the experiments.\loosepar{}

\myparagraph{Evaluation protocol. } As we describe in the experimental setup, the videos naturally include various visual backgrounds and camera perspectives that are significantly different from the robot workspace. Therefore, we only intentionally vary two dimensions before evaluating each trial of robot execution, namely the spatial layouts and the new object instances. 
Furthermore, the new object generalizations are included in the tasks \mugcoaster{}, \chip{}, \juice{}, \peasplate{}, and \cheeseplate{}. As for the other tasks, there are no novel objects involved, but we extensively vary the spatial layouts of task-relevant objects for evaluation. The policy performance of a task is the average success rate over 15 real-world trials. Aside from the success rates, we also group the failed executions into three types: \textit{Missed tracking} of objects due to failure of the vision models, \textit{Missed grasping} of objects during execution, and \textit{Unsatisfied contacts} where the target object configurations are not achieved for reasons other than the previous two failure types.

\begin{figure}
    \centering
    \includegraphics[width=\linewidth]{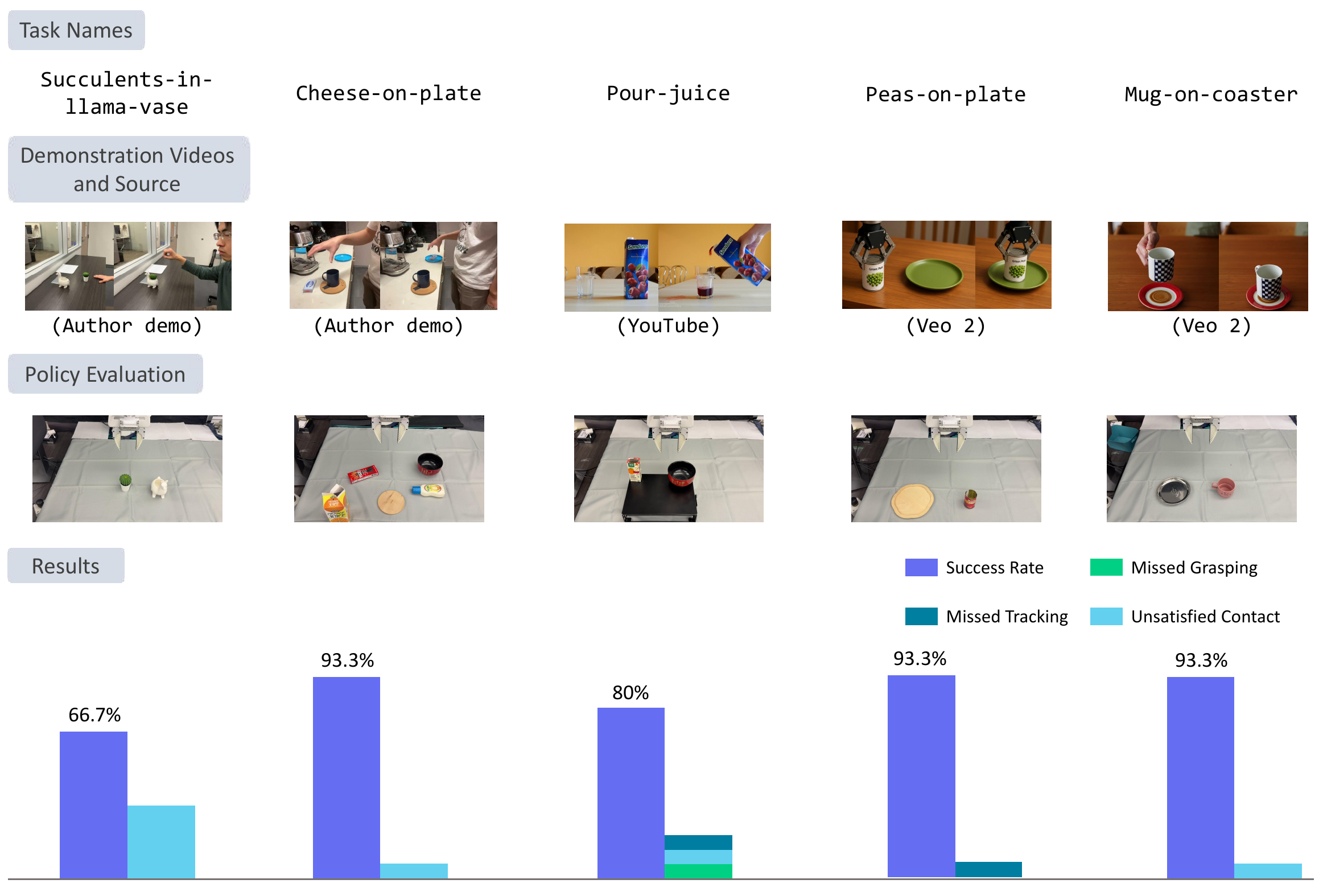}
    \caption{This figure shows results in the same format as Figure \ref{fig:tasks}, with the difference that demonstration videos do not contain depth information and that videos come from different sources. Note that the \mugcoaster{} task uses a different demonstration video to showcase compatibility with different object instances, camera angles, and video sources.}
    \label{fig:tasks_rgb}
\end{figure}

\myparagraph{Baselines. } To understand the model capacity and validate our design choices, we compare \ourmethod{} with baselines. Since no prior work exists \footnote{See Section \ref{related:close_works} for detailed discussion on related work} that matches the exact setting of our approach, we adopt the most important components from prior works and treat them as baselines to our model. Specifically, we implement the following two baselines: 1) \textsc{Hand-motion-imitation}~\cite{wang2023mimicplay, bharadhwaj2023zero} is a baseline that predicts robot actions by learning from the hand trajectories. The rest of the parts remain the same as \ourmethod{}. We use this baseline to show whether it is critical to compute actions centered around objects. 2) \textsc{Dense-Correspondence}~\cite{ko2023learning, heppert2024ditto} is a baseline that replace the TAP model in \ourmethod{} with a dense correspondence model, optical flows. This baseline is used to evaluate whether our choice of the TAP model is a better design. For this ablative study, we conduct experiments on \mugcoaster{} and \simpleboat{} to validate our model design, covering the distribution of common daily objects and assembly manipulation that requires precise control.

\begin{figure}[t]
\centering
\begin{minipage}{\linewidth}
\centering
\includegraphics[width=\linewidth]{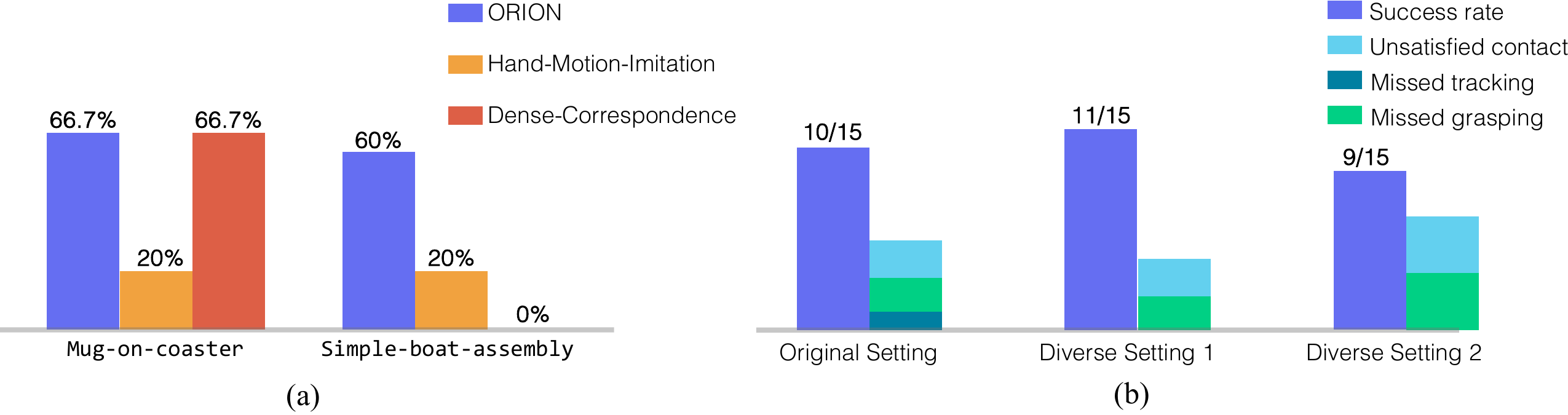}
\caption{(a) Experimental comparison between \ourmethod{} and the two baselines, namely \textsc{Hand-Motion-Imitation} and \textsc{Dense-Correspondence}. (b) Ablation study on using different videos of the same task. We select the task \mugcoaster{} for conducting this ablation. We display the number of successful trials out of 15 total trials on the bar plots for each setting. Figure~\ref{fig:supp-video} in Appendix~\ref{supp:video-diverse} visualizes the different settings in this experiment.}
\vspace{-0.5cm}
\label{fig:ablation}
\end{minipage}
\end{figure}

\subsection{Experimental Results}
\label{sec:exp-quant}

Our evaluations are presented in Figures~\ref{fig:tasks}
~\ref{fig:tasks_rgb}, and ~\ref{fig:ablation}.
We answer question (1) by showing the successful deployment of the \ourmethod{} policies, while no other methods are designed to be able to operate in our setting. Furthermore, \ourmethod{} yields an average of $66.7\%$ success rate, which validates our model design in learning from a single human video in the open-world setting.

We then answer question (2), showing the comparison results in Figure~\ref{fig:ablation} against the baseline, \textsc{Hand-motion-imitation}, which yields low success rates in both tasks. Concretely, \textsc{Hand-motion-imitation} typically succeeds in trials where the initial spatial layouts are similar to the one in $V$. Its major failure mode is not being able to reach the target object configuration, e.g., misplacing the mug on the table while not achieving contact with the coaster. These results imply that learning from human hand motion from $V$ results in poor generalization abilities of policies, supporting the design choice of \ourmethod{} which focuses on the object-centric information. \loosepar{}

We further answer question (3) by comparing the performance between \ourmethod{} and the baseline, \textsc{Dense-Correspondence}, shown in Figure~\ref{fig:ablation}(a). We observe that the optical flow baseline performs drastically worse on \simpleboat{} than on \mugcoaster{}. With our further investigation, we find that the optical flow baseline discovers keyframes in the middle of smooth transitions as opposed to changes in object contact relations, resulting in a manipulation plan that computes completely wrong actions. This finding further supports our choice of using TAP keypoints to discover the keyframes. \loosepar{}

To answer question (4), we conduct controlled experiments using the task \mugcoaster{}. Specifically, we record two additional videos of the same task in very different visual conditions and spatial layouts (see details in Appendix~\ref{supp:video-diverse}) and construct a policy from each video. Then, we compare the two policies against the original one and test them using the same set of evaluation conditions. The results in Figure~\ref{fig:ablation}(b) show that there is no statistically significant difference in the performance, demonstrating that \ourmethod{} is robust to videos taken under very different visual conditions. Next, to answer question (5), we show that \ourmethod{} is effective in scaling to long-horizon tasks. This conclusion is supported by the performance among the pairs of \mugcoaster{} versus \rearrange{}, and \simpleboat{} versus \complexboat{}. In these two pairs, both the short-horizon tasks are subgoals of their long-horizon counterparts, yet we do not see any performance drop between the two. This result indicates that \ourmethod{} excels at scaling to long-horizon tasks without a significant drop in policy performance. \loosepar{}

Finally, we answer question (6) by observing a similar successful deployment of \ourmethod{}'s policies when demonstration videos only contain RGB information without depth, as shown in Figure~\ref{fig:tasks_rgb}. \juice{} is retrieved from an in-the-wild YouTube video, \peasplate{} and \mugcoaster{} are generated from Veo 2 \cite{veo2}, and \llama{} and \cheeseplate{} are from the same recordings used for RGB-D experiments but with the depth channel removed. We observe an average success rate of $85.3\%$, showing that our method can perform well even without depth information. In addition, we see that our method is able to handle demonstration videos from different sources (from controlled environments, Internet videos, and generated models) and different camera angles, along with test-time object instances and camera angles that differ from the demonstration. These results support our conclusions for question (4) in the RGB-only setting. To directly compare the RGB-only and RGB-D pipelines, we inspect the results of the \llama{} task, on which both pipelines were evaluated. Interestingly, we see that the RGB-only pipeline performs comparably and even slightly better than the RGB-D pipeline on the same task. We note that certain RGB-only architectural changes, such as the shift from 3D keypoint trajectories to 6-DOF pose trajectories, allow the model to trade noise in global registration and end-effector trajectory optimization for noise in object pose estimation and mesh generation models. We observe that the strong world knowledge embedded in these models allows our RGB-only pipeline to perform competitively with our RGB-D pipeline, compensating for the lack of depth with minimal added noise.

%% file: 02_related.tex
\myparagraph{Learning Manipulation From Human Videos. } 
Human videos offer a rich repertoire of object interaction behaviors, making them an invaluable data source for manipulation. A large body of work has explored how to leverage human video data for learning robot manipulation~\cite{wang2023mimicplay, xiong2021learning, bahl2022human, kumar2023graph, liu2018imitation, sharma2019third, smith2019avid, xu2023xskill}, either through pre-training a single latent representation~\cite{nair2022r3m, wang2023mimicplay, xu2023xskill}, learning an implicit reward function~\cite{chen2021learning, ma2022vip}, or learning generative models that in-paint human morphologies~\cite{ko2023learning, bharadhwaj2023zero, bahl2022human, bharadhwaj2023towards}. However, they either require additional robot data from the target tasks or paired data between humans and robots. Our approach takes a different direction by tackling how a robot can imitate or learn from a single human video only: the robot does not rely on pre-existing data, models, or ground-truth annotations \textit{in scenes} where video recording and robot evaluation take place. We refer to such a setting as \textit{open-world imitation from observation}, where the robot is not programmed or trained to interact with the objects in the video \textit{a priori} and the video data does not come with any robot actions. Our setting is closely related to the problem of ``Imitation Learning from Observation''~\cite{torabi2021imitation}, where state-only demonstrations are used to construct policies for physical interaction. However, this line of prior work assumes simulators of demonstrated tasks exist and physical states of the agents or objects are known~\cite{pavse2020ridm, karnan2022adversarial, torabi2019imitation, torabi2018generative, torabi2018behavioral}. In contrast, our setting does not assume the digital replica of real-world tasks, and all the object information is perceived only through videos. 

\myparagraph{One-Shot Imitation Learning for Manipulation. } 
Studies have delved into learning manipulation policies from one demonstration. A notable frame is one-shot imitation learning within a meta-learning framework proposed by Duan et al.~\cite{duan2017one}. While prior works on one-shot imitation learning have shown a robot performing new tasks from one demonstration, they require extensive in-domain data and a well-curated set of meta-training tasks beforehand, leading to significant data collection costs and restricted policy generalization at test time due to the tailored nature of the training.\loosepar{}

An alternative approach involves using a single demonstration for initial guidance, refining the policy through real-world self-play~\cite{di2022learning, haldar2023watch,haldar2023teach, johns2021coarse, valassakis2022demonstrate}. 
However, this approach mainly applies to reset-free tasks and struggles with scaling to multi-stage tasks where resetting to the task's initial conditions does not come free. Our work aligns with these studies in using a single demonstration for learning manipulation, but stands out by not needing prior data or self-play. With just one single human video, our method constructs a policy that successfully completes the task, adapting to various visual and spatial differences from the task instance of video demonstration.

\label{related:close_works}
\myparagraph{Open-World Imitation Learning in Robotics. } 
Recent work \cite{patel2025robotic, park2025demodiffusion} has begun to bridge single-demonstration learning and learning from human videos, closely matching our ``open-world" setting. Similarly to \ourmethod{}, these works utilize the broad world knowledge of off-the-shelf models. However, they typically rely on stronger geometric inputs (\textit{e.g.}, calibrated RGB-D or multi-view captures, explicit 3D hand poses, or pre-scanned object meshes) that are rarely available in casual consumer or Internet videos. These dependencies limit applicability to uncontrolled, single-view recordings. In contrast, our method requires only the demonstration video (RGB, or optionally RGB-D) with no pre-scanned object models. 

Another recent approach, OKAMI \cite{li2024okamiteachinghumanoidrobots}, most directly addresses the problem of ``open-world imitation from observation". However, OKAMI and \ourmethod{} represent fundamentally different approaches to imitation. OKAMI's method is \textit{demonstrator-centric}, focusing on reconstructing the demonstrator's body and hand motions and retargeting them onto a humanoid robot. In contrast, our method is \textit{object-centric}, focusing on generating a plan that replicates the observed object trajectories, independent of the human's specific pose. In addition, this means that \ourmethod{} does not need a demonstration video showing the human's entire body. By extracting object-centric Open-world Object Graphs (OOGs) with off-the-shelf vision foundation models, \ourmethod{} recovers task-relevant object states and interaction cues directly from the video and constructs robot-executable plans that generalize across backgrounds, viewpoints, spatial layouts, and novel object instances.

\myparagraph{Object-Centric Representations for Robot Manipulation. }
The concept of object-centric representation has long been recognized for its potential to enhance robotic perception and manipulation by focusing on the objects within a scene. Prior works have shown the effectiveness of such representation in downstream manipulation tasks by factorizing visual scenes into disentangled object concepts~\cite{tremblay2018deep, tyree20226, migimatsu2020object, wang2019deep, devin2018deep}, but these works are typically confined to known object categories or instances. Recent developments in foundation models allow robots to access the open-world object concepts through pre-trained vision models~\cite{kirillov2023segment, oquab2023dinov2}, enabling a wide range of abilities such as imitation of long-horizon tabletop manipulation~\cite{zhu2022viola, shiplug} or mobile manipulation in the wild~\cite{stone2023open}. Building upon these advances, our work focuses on leveraging open-world, object-centric concepts in imitating manipulation behaviors from actionless human videos. We propose a graph-based representation called Open-world Object Graph (OOG), which allows a robot to imitate from a human video by leveraging the object-centric concepts. This proposed representation shares a similar vein with prior works that factorize scene or task-relevant visual concepts into scene graphs~\cite{kumar2023graph, mo2019structurenet, huang2023planning, qureshi2021nerp, zhu2021hierarchical}. However, our representation is tailored to integrate open-world object concepts and enable generalization across different embodiments, specifically a human and a robot.\loosepar{}
\vspace{-0.2cm}

%% file: 06_conclusions.tex
\vspace{-0.2cm}
In this paper, we investigate the problem of learning robot manipulation from a single human video in the open-world setting, where a robot must learn to manipulate novel objects using one video demonstration. To tackle this problem, we introduce \ourmethod{}, an algorithm built on object-centric priors. Our results show that given a single human video, with or without depth, \ourmethod{} is able to construct a policy that generalizes over the following four dimensions: visual backgrounds, camera angles, spatial layouts, and the presence of new object instances. 

Limitations: We consider task goals to be described by contact states, thereby naturally avoiding the ambiguities introduced when considering spatial relations, such as placing items next to an object. Inferring human intentions while clearing the inherent ambiguities in videos is a future direction to explore. We have also assumed that demonstration videos must be captured with a stationary camera. In reality, most videos in everyday scenarios contain a moving camera. 
Thus, a promising future direction is to investigate how to build a model that can reconstruct dynamic scenes from a moving camera, where the desired model can estimate the geometry of both static and moving objects in the scenes while ensuring the scale of the reconstructed scenes and objects matches the real world. As vision models continue to improve, stronger perceptual back-ends can be integrated into our framework with minimal modification, further boosting success rates and enabling a broader range of tasks. 


%% file: supp.tex
\section{Additional Technical Details}
\subsection{Data Structure of an OOG.} For easy reproducibility of the proposed method, we present a table that explains the data structure of an OOG.

\begin{table}[htbp]
    \centering
    \begin{tabular}{@{} m{3cm} m{3cm} m{7cm} @{}}
        \toprule
        \textbf{Node/Edge} & \textbf{Type} & \textbf{Attributes} \\
        \midrule
        $\G.vo_i$ & Object Node & 3D point cloud of an object (+ object mesh if RGB-only). \\    
        $\G.vh$ & Grasp Node & Grasp locations on object, calculated through hand mesh finger locations / object mesh heuristics. \\
        $\G.vp_{ij}$ & Point Node & A trajectory of a TAP keypoint (RGB-D) or 6-DOF object pose (RGB-only) between two keyframes. \\
        
        \midrule
        $\G.eo_{ik}$ & Object-Object Edge & A binary value of contact or not. \\
        $\G.eh_{i}$ & Object-Hand Edge & A binary value of contact or not. \\
        $\G.ep_{ij}$ & Object-Point Edge & The presence of an edge represents the belonging relation, and no specific feature is attached.\\
    \bottomrule
    \vspace{0.5cm}
    \end{tabular}
    \caption{Data Structure of an OOG. For a given OOG $\G=(\mathcal{V}, \mathcal{E})$, it has $\mathcal{V}=\{\G.vo_i\} \cup \{\G.vh\} \cup \{\G.vp_{ij}\}$, and $\mathcal{E}=\{\G.eo_{ik}\} \cup \{\G.eh_{i}\} \cup \{\G.ep_{ij}\}$.
    }
    \label{tab:oog_data_structure}
\end{table}

\subsection{Implementation Details}
\label{supp:implementation}

\myparagraph{Changepoint detections.} 
We use changepoint detection to identify changes in velocity statistics of TAP keypoints. Specifically, we use a kernel-based changepoint detection method and choose radial basis function~\cite{killick2012optimal}. The implementation of this function is directly based on an existing library Ruptures~\cite{truong2020selective}. 

\myparagraph{Plane estimation.} In Section~\ref{sec:plan-generation}, we mentioned using the prior knowledge of tabletop manipulation scenarios and transforming the point clouds by estimating the table plane. Here, we explain how the plane estimation is computed. Concretely, we rely on the plane estimation function from Open3D~\cite{zhou2018open3d}, which gives an equation in the form of $ax + by + cz = d$. From this estimated plane equation, we can infer a normal vector of the estimated table plane, $(a, b, c)$, in the camera coordinate frame. Then, we align this plane with xy plane in the world coordinate frame, where we compute a transformation matrix that displaces the normal vector $(a, b, c)$ to the normalized vector $(0, 0, 1)$ along the z-axis of the world coordinate frame. This transformation matrix is used to transform point clouds in every frame so that the plane of the table always aligns with the xy plane of the world coordinate.

\myparagraph{Object localization at test time.} When we localize objects at test time, there could be some false positive segmentation of distracting objects. Such vision failures will prevent the robot policy from successfully executing actions. To exclude such false positive object segmentationt, we use Segmentation Correspondence Model (SCM) from GROOT~\cite{zhu2023groot}, where SCM filters out the false positive segmentation of the objects by computing the affinity scores between masks using DINOv2 features. 

\myparagraph{Global registration.} In this paper, we use global registration to compute the transformation between observed object point clouds from videos and those from rollout settings. We implement this part using a RANSAC-based registration function from Open3D~\cite{zhou2018open3d}. Specifically, given two object point clouds, we first compute their features using Fast-Point Feature Histograms (FPFH)~\cite{rusu2009fast}, and then perform a global RANSAC registration on the FPFH features of the point clouds~\cite{choi2015robust}. 

\myparagraph{Implementation of SE(3) optimization.} 
We parameterize each homogeneous matrix $T_{i}$ into a translation variable and a rotation variable and randomly initialize each variable using the normal distribution. We choose quaternions as the representation for rotation variables, and we normalize the randomly initialized vectors for rotation so that they remain unit quaternions. 
With such parameterization, we optimize the SE(3) end-effector trajectories $T_0, T_1, \dots, T_{t_{l+1}-t_{l}}$ over the Objective (\ref{label:optim}). However, jointly optimizing both translation and rotation from scratch typically results in trivial solutions, where the rotation variables do not change much from the initialization due to the vanishing gradients. To avoid trivial solutions, we implement a two-stage process. In the first stage, we only optimize the rotation variables with 200 gradient steps. Then, the optimization proceeds to the second stage, where we optimize both the rotation and translation variables for another 200 gradient steps. In this case, we prevent the optimization process from getting stuck in trivial solutions for rotation variables. We implement the optimization process using Lietorch~\cite{teed2021tangent}. 


\myparagraph{FoundationPose implementation. } The off-the-shelf FoundationPose implementation does not support RGB-only inputs. To adapt it for our use case, we modify the implementation to take in a tuple $(l,h)$ where $l$ and $h$ represent the lower and upper bounds of the distance from the camera to the object, respectively. We estimate this upper and lower bound by applying monocular depth estimation on the object, calculating the median depth over all pixels belonging to the object, and setting $l=med - 0.15$ and $h=med + 0.15$. We then create pose hypotheses centered at regular intervals between $l$ and $h$. These pose hypotheses are then sent to the pose refiner, and from here, the implementation follows the original FoundationPose.

\myparagraph{Scaling object meshes. } The meshes output by InstantMesh are of arbitrary scale. To deal with this, we use the estimated depth from monocular depth estimation models to backproject each pixel within the object mask, creating a point cloud of that object. To remove outliers in the point cloud, we use Open3D's remove\_statistical\_outlier and cluster\_dbscan methods. Next, we find the axis-aligned bounding box of the point cloud and mesh. We calculate a scale factor $s=\dfrac{p}{m}$ where $p$ and $m$ are the diagonal lengths of the point-cloud and mesh bounding boxes respectively. Finally, we uniformly apply $s$ to the generated mesh to get an approximation of the true mesh scale.

\myparagraph{\ourmethod{} RGB vs RGB-D implementation.}
Below we summarize the implementation-level differences between our RGB-D and RGB-only pipelines. These differences affected how object correspondence and end-effector (EE) trajectories are obtained, and which assumptions are introduced in the pipeline

\begin{itemize}
  \item \textbf{Tracking and transform estimation.} Our RGB-D pipeline tracks 3D keypoint trajectories on objects and uses \emph{global registration} of object point clouds to estimate the correspondence transform between demonstration and rollout scenes:
  \[
    T_{\hat{o}\rightarrow o} \;=\; \operatorname{GLOBAL\_REGISTRATION}(\hat{o}, o),
  \]
  where $\hat{o}$ and $o$ are object point clouds of the demonstration and rollout objects, respectively.

The RGB-only pipeline directly estimates the full 6-DOF pose of the target object in each frame during the demonstration, along with the initial pose of the target and reference objects during rollout. Thus, we only need to find the desired end pose of the target object during rollout. In this case, $o \text{ and } \hat{o}$ refers to the object's pose. We find the rollout target end pose with  \[
    {o}_{target}^{end} = o_{ref}^{start} \cdot (\hat{o}_{ref}^{start})^{-1} \cdot \hat{o}_{target}^{start}
  \]
  removing the need for an intermediate point-cloud registration step
  \item \textbf{End-effector (EE) pose sequence.} The RGB-D pipeline finds the EE sequence by solving the optimization problem in \ref{label:optim}. However, in the RGB-only case, because we already have the pose sequence of the object, we can apply a single transform to all elements of $o_{target}$ to retrieve $\{T\}$ 
  \item \textbf{Grasp selection.} The RGB-only pipeline chooses grasp locations using object geometry heuristics derived from estimated meshes, rather than attempting to recover demonstrated hand/grasps from video. This is an assumption that simplifies grasp selection but can bias results relative to the RGB-D pipeline, which attempts to infer demonstrated hand information. In addition, the pose of this grasp selection is the transform applied to all elements of $o_{target}$ to retrieve the EE pose trajectory. 
\end{itemize}

\myparagraph{Object‑Centric Trajectory Warping}

\begin{algorithm}[h]
\caption{Translation Warping}
\label{alg:trans-warp}
\begin{algorithmic}[1]
  \Require 3D demonstration object trajectory $\tau$, Desired 3D new start and end points $s, e$. 
  \Ensure Warped 3D object trajectory $\tau^{warped}$
    \Statex \textbf{// Normalize trajectory and calculate rotation with Rodrigues formula}
    \State $\tau^{norm} \leftarrow \{\tau_i - \tau_{1} \ \forall \ i \in \{1, ..., L\}\}$
    \State $v_1, v_2 \leftarrow \tau_L - \tau_1, e - s$
    \Statex \textbf{// Calculate rotation with Rodrigues formula}
    \State $A \leftarrow $ \Call{Rodrigues}{$v_1,v_2$}
    \State $\mathit{scale} \leftarrow \dfrac{\norm{v_2}}{\norm{v_1}}$
    \Statex \textbf{// Transform back to workspace frame}
    \State $\tau^{warped} \leftarrow \{ \mathit{scale}( A \cdot \tau^{norm}) + s \}$
    \State 
  \State \Return $\tau^{warped}$
\end{algorithmic}
\end{algorithm}

\begin{algorithm}[h]
\caption{Rotation Warping}
\label{alg:rot-warp}
\begin{algorithmic}[1]
  \Require 6-DOF demonstration object trajectory $\tau$, desired 6-DOF start $s$ and end $e$. For some 4x4 6-DOF pose matrix $X$, we refer to the rotation as $Rot(X) \leftarrow X_{1:3,1:3}$ and translation as $t(X) \leftarrow X_{1:3,4}$. 
  \Ensure Warped object rotation trajectory $R^{warped}$
    \State $R \leftarrow Rot(\tau)$
    \State $v_1, v_2 \leftarrow t(\tau_L) - t(\tau_1), t(e) - t(s)$
    \Statex \textbf{// Rotate $R$ to align with demonstration direction }
    \State $A \leftarrow $ \Call{Rodrigues}{$v_1,v_2$} 
    \State $R' \leftarrow \{ A \cdot R_i \ \forall \ i \in \{1, ..., L\}\}$
    \State ${R'}^{norm} \leftarrow \{ R'_i \cdot (R'_1)^T \ \forall \ i\}$ 
    \Statex \textbf{// Calculate boundary rotations and apply spherical linear interpolation}
    \State $\mathit{slerp_{start}} \leftarrow Rot(s), \mathit{slerp_{end}} \leftarrow ({{R'}^{norm}_N})^T \cdot Rot(e)$
    \State $I \leftarrow SLERP(slerp_{start}, slerp_{end})$ 
    \Statex \textbf{// Combine interpolated rotations with original trajectory}
    \State $R^{warped} \leftarrow \{ {R'}^{norm}_i\cdot I_i \ \forall \ i \}$ 
  \State \Return $R^{warped}$
\end{algorithmic}
\end{algorithm}

\section{System Setup}
\myparagraph{Details of camera observations.} As mentioned in Section~\ref{sec:experiments}, we use an iPad with a TrueDepth camera for collecting RGB-D human video demonstrations. We use an iOS app, Record3D, that allows us to access the depth images from the TrueDepth camera. We record RGB and depth image frames in sizes $1920\times 1080$ and $640\times 480$, respectively. To align the RGB images with the depth data, we resize the RGB frames to the size $640\times 480$. The app also automatically records the camera intrinsics of the iPhone camera so that the back-projection of point clouds is made possible. In the RGB-only case, we use the same videos with the depth channel removed, videos generated from generative models, and YouTube videos.

To stream images at test time, we use an Intel Realsense D435i. In our robot experiments, we use RGB and depth images in the size $640\times 480$ or $1280\times720$ in varied scenarios, all covered in our evaluations. Evaluating on different image sizes showcases that our method is not tailored to specific camera configurations, supporting the wide applicability of constructed policy.

\myparagraph{Implementation of real robot control.} In our evaluation, we reset the robot to a default joint position before object interaction every time. Then we use a reaching primitive for the robot to reach the interaction points. Resetting to the default joint position enables an unoccluded observation of task-relevant objects at the start of each decision-making step. Note that the execution of object interaction does not necessarily require resetting. To command the robot to interact with objects, we convert the optimized SE(3) action sequence to a sequence of joint configurations using inverse kinematics and control the robot using joint impedance control. We use the implementation of Deoxys~\cite{zhu2022viola} for the joint impedance controller that operates at $500$ Hz. To avoid abrupt motion and make sure the actions are smooth, we further interpolate the joint sequence from the result of inverse kinematics. Specifically, we choose the interpolation so that the maximal displacement for each joint does not exceed $0.5$ radian between two adjacent waypoints. 

\section{Success conditions of tasks}
\label{supp:success}
We describe the success conditions for each of the tasks in detail: 
\begin{itemize}
    \item \mugcoaster{}: A mug is placed upright on the coaster.
    \item \simpleboat{}: A red block is placed in the slot closest to the back of the boat. The block needs to be upright in the slot.  
    \item \chip{}: A bag of chips is placed on the plate, and the bag does not touch the table. 
    \item \llama{}: A pot of succulents is inserted into a white vase in the shape of a llama. 
    \item \rearrange{}: The mug is placed upright on the coaster, and the cream cheese box is placed on the plate.
    \item \complexboat{}: The chimney-like part is placed in the slot closest to the front of the boat. The red block is placed in the slot closest to the back of the boat. Both blocks need to be upright in the slots. 
    \item \breakfast{}: The mug is placed on top of a coaster, the cream cheese box is placed in the large area of the plate, and the food can is placed on the small area as shown in the video demonstration. 
    \item \peasplate{}: The peas can is placed upright in the large area of the plate.
    \item \juice{}: The juice box is tilted 45 degrees in the direction of its opening, such that the opening ends above a container.
    \item \cheeseplate{}: The cream cheese box is placed in the large area of the plate.
\end{itemize}

In practice, we record the success and failure of a rollout as follows: If the program in \ourmethod{} policy returns true when matching the observed state with the final OOG from a plan, we mark a trial as success as long as we observe that the object state indeed satisfies the success condition of a task as described above. Otherwise, if the robot generates dangerous actions (bumping into the table) or does not achieve the desired subgoal after executing the computed trajectory, we consider the rollout as a failure and we manually record the failure.

\section{Additional Details on Experiments}
\label{supp:video-diverse}

\begin{figure*}[htp]
    \centering
    \includegraphics[width=1.0\linewidth, trim=0cm 0cm 0cm 0cm,clip]{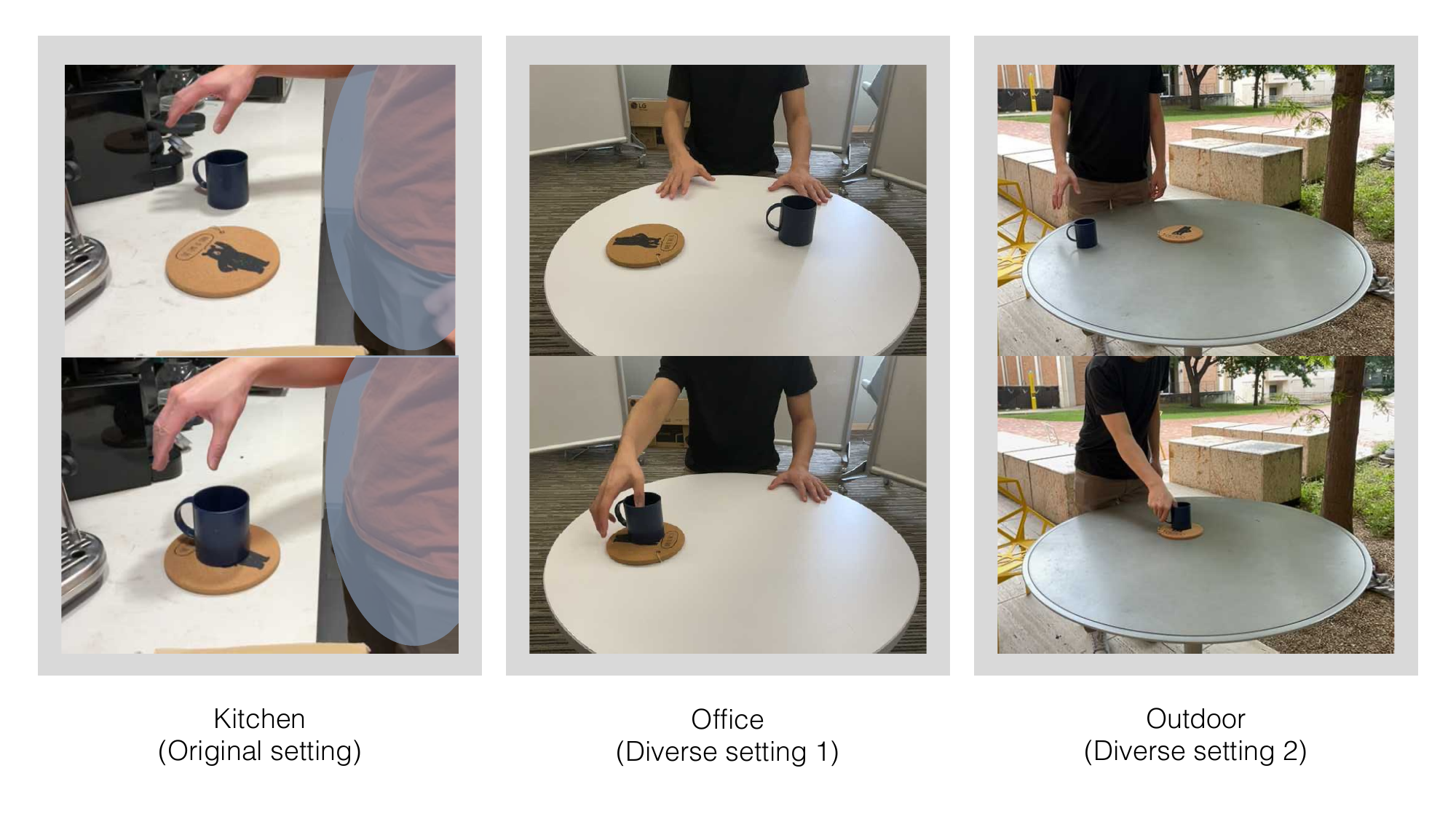}
    \caption{This figure visualizes the initial and final frames of the three videos of the same task~\mugcoaster{}.}
    \label{fig:supp-video}
\end{figure*}

\myparagraph{Diverse video recordings used in the ablation study.} Figure~\ref{fig:supp-video} shows the three videos taken in very different scenarios: kitchen, office, and outdoor. The video taken in kitchen scenario is used in the major quantitative evaluation, termed ``Original setting''. The other two settings are termed ``Diverse setting 1'' and ``Diverse setting 2.'' We conduct an ablation study where we compare policies imitated from these three videos, which inherently involve varied visual scenes, camera perspectives. The result of the ablation study is shown in Figure~\ref{fig:ablation}.

\section{Prompts sent to VLM}
\label{supp:prompt}
\myparagraph{Identify task-relevant objects during rollout. } This prompt is used during to identify which objects in the scene correspond to the task-relevant objects in the demonstration by comparing the object similarity.
\begin{formal}
\textbf{Prompt:} You are given a list of object names (called obj\_names): \{obj\_names\}, a list of object types (called obj\_types): \{obj\_types\}, and images with white text at the bottom. One image is the current observation in real life, which is labeled accordingly. Another image is the observation from the demonstration video, which annotates the objects in the video according to object\_names. The rest of the images, which are labeled as "Demo Video Frame x", show how the object\_names move in the demonstration video. obj\_types[i] describes what type of object object\_names[i] is. You can see the corresponding object in the image titled "Demo Observation Image". If the object type is "manipulate", that means a person manipulated the object with their hands. If the object type is "reference", that means the object was used as a referece for the manipulate object. For example, if a person shakes salt from a salt shaker into a bowl, the salt shaker is the manipulate object and the bowl is the reference object, since the person directly manipulated the salt shaker, and the bowl stayed still and received the salt from the salt shaker. For each object name in obj\_names, return an object name (can be the same name) that best describes the closest related object in the image titled "Real Observation Image". If the closest related object in the Real Observation Image is exactly the same object as the one in the Demo Observation Image, you can return the same name. Make sure that obj\_names[i] corresponds to similar\_obj\_names[i], and similar\_obj\_names[i] is of type obj\_types[i]. Finally, please make sure the similar\_obj\_names[i] is described thoroughly, including color or shape. In descending order of priority, similar\_obj\_names[i] should have a similar shape, object type, orientation or function to obj\_names[i]. Make sure each description is descriptive, but concise (5 words or less).

A description of the video from which the demonstration frames were taken is "{demo\_video\_description}"

Your output format is:
\begin{verbatim}
{{
"similar_obj_names": [NAME_OF_SIMILAR_OBJ_1,
                    NAME_OF_SIMILAR_OBJ_2, ...]
"explanation": "Here, you will explain your reasoning for 
                choosing similar_obj_names[i] as the 
                similar object to obj_names[i]" 
}}
\end{verbatim}

Example:
\begin{verbatim}
{{
"similar_obj_names": [red water bottle, checkered plate]
"explanation": "obj_names[0] is 'green thermos', and the 
closest related object in the image is a red water bottle, 
because they are both drinking objects which are both standing 
upright, are both cylindrical, and both around 0.3 meters. 
obj_names[1] is 'purple plate', and the closest related 
object in the image is a checkered plate, which has the 
same shape and function as the purple plate, and can be 
used as a reference object for the red water bottle. 
Taking the whole real-world scene into consideration, 
we can place the red water bottle onto the checkered 
plate, just as the green thermos is placed onto 
the purple plate in the demonstration video."
}}
\end{verbatim}

Ensure the response can be parsed by Python 'json.loads', e.g.: no trailing commas, no single quotes, etc.
\end{formal}

\myparagraph{Identify task relevant objects, then classify into target and reference objects. } The prompts used here are largely similar to the ones used in OKAMI \cite{li2024okamiteachinghumanoidrobots}.

\begin{formal}
\textbf{Find task relevant objects prompt:} You need to analyze what the human is doing in the images, then tell me: 1. All the objects in front of the scene (mostly on a plane similar to a table). You should ignore the background objects. 2. The objects of interest. They should be a subset of your answer to the first question. They are likely the objects manipulated by the human or objects that interact with other objects. Objects that the human uses (like utensils) to directly manipulate other objects should NOT be included as task relevant objects. Note that there are irrelevant objects in the scene, such as objects that are not interacted with by the human and/or the other objects. You should ignore the irelevant objects. Make sure to include a color and texture description of the relevant objects. This is especially important if there are many instances of a single type of object. For example, if there are blue, green, and yellow balls, be sure to specify each one individually. Do NOT just say "colorful balls". Finally, if any objects emit liquid, do not include the liquid itself as an object you return. Your output format is:

\begin{verbatim}
{
    "video_description": "The human is xxx.",
    "all_objects": ["OBJECT1", "OBJECT2", ...], 
    "objects_of_interest": ["OBJECT1", "OBJECT3", ...],
}    
\end{verbatim}

Ensure the response can be parsed by Python 'json.loads', e.g.: no trailing commas, no single quotes, etc.

Example:
\begin{verbatim}
{
    "video_description": "The human is picking up
    a red ball with blue dots and putting it in
    a green cup",
    "all_objects": ["blue ball with stripes",
                    "small green ball", 
                    "red ball with blue dots", 
                    "green cup"], 
    "objects_of_interest": ["red ball with blue dots", 
                            "green cup"],
}
\end{verbatim}
\end{formal}

\begin{formal}
\textbf{Identify target objects prompt:} The following images shows a manipulation motion, where the human is manipulating an object or objects. 

Your task is to determine which objects are being manipulated in the images below. You need to choose from the following objects: \{task\_relevant\_object\_candidates\}.

Tips: A manipulated object is an object that the human is interacting with, such as picking up, moving, or pressing, and it is in contact with the human or an object (like a utensil) that the human is using. If there are multiple objects of the same type, be sure to specify the color, texture, etc. that correspond to the manipulated object

Your output format is:
\begin{verbatim}
{{
"manipulate_object_names": [
            "MANIPULATE_OBJECT_DESCRIPTIVE_NAME_1", 
            "MANIPULATE_OBJECT_DESCRIPTIVE_NAME_2", 
            ...],
}}    
\end{verbatim}

Example:
\begin{verbatim}
{{
    "manipulate_object_names": ["red ball with blue dots", 
                                "green cup"],
}}

\end{verbatim}
Ensure the response can be parsed by Python 'json.loads', e.g.: no trailing commas, no single quotes, etc.
\end{formal}

\begin{formal}
\textbf{Identify reference objects prompt:} The following images shows a manipulation motion, where the human is manipulating the objects in the list - \{manipulate\_obj\_names\}.

Please identify the reference object in the image below, which could be an object on which to place an object in the list \{manipulate\_obj\_names\}, or an object that something in the list \{manipulate\_obj\_names\} is interacting with. You need to first identify whether there is a reference object. If so, you need to output the reference object's name chosen from the following objects: \{task\_relevant\_object\_candidates\}.
Note that there may not necessarily be a reference object. Examples of reference objects are: a coaster, if the motion involves picking up a cup and placing it on a coaster; a bowl, if the motion involves picking up a bottle and pouring it into the bowl.

Finally, make sure your proposed ith reference object corresponds to the given ith manipulated object.

\begin{verbatim}
{{
"reference_object_names": ["REFERENCE_OBJECT_NAME_1",
                          "REFERENCE_OBJECT_NAME_2", ...]
                            or ["None"],
"reasoning": "Here, you will explain your reasoning for 
choosing these objects as the reference object, 
or for saying that there are no reference objects"
}}    
\end{verbatim}
Your output format is:

Example:
\begin{verbatim}
{{
    "reference_object_names": ["white circular plate",
                                "brown coaster"],
    "reasoning": The white circular plate is a reference 
    object because the human is placing the red ball on it. 
    The brown coaster is a reference object because 
    the human is placing the green cup on it.
}}

\end{verbatim}
Ensure the response can be parsed by Python 'json.loads', e.g.: no trailing commas, no single
quotes, etc.
\end{formal}